# DikpolaSat Mission:
## Improvement of Space Flight Performance and Optimal Control Using Trained Deep Neural Network - Trajectory Controller for Space Objects Collision Avoidance

Manuel Ntumba [1], Saurabh Gore [2], Jean-Baptiste Awanyo [3]

[1][2][3] *Division of Space Applications, Tod'Aérs - Transparent Organization for Aeronautics and Space Research, Lomé, Région Maritime, Togo.*

[1] *Space Generation Advisory Council, Vienna, Austria.*

[2] *Moscow Aviation Institute, Moscow, Russia.*

[3] *Univerité Sultan Moulay Slimane, Béni Mellal, Morocco.*

manuel.ntumba@spacegeneration.org

sdgore@mai.education

**ABSTRACT**

This paper introduced the space mission *DikpolaSat Mission*, how this research fits into the mission, and the importance of having a trained DNN model instead of the usual GN&C functionality. This paper shows how the controller demonstration is carried out by having the spacecraft follow a desired path, specified in the referenced model. Increases can be made by examining the route used to construct a DNN and understanding the effects of various activating functions on system efficiency. The obstacle avoidance algorithm is built into the control features to respond spontaneously using inputs from the neural network for collision avoidance while optimizing the modified trajectory. The action of a neural network to control the adaptive nature of the nonlinear mechanisms in the controller will make the control system capable of handling multiple nonlinear events and also uncertainties that have not been induced in the control algorithm. Multiple algorithms for optimizing flight controls and fuel consumption can be implemented using knowledge of flight dynamics in trajectory and also in the event of obstacle avoidance. This paper also explains how a DNN can learn to control the flight path and make the system more reliable with each launch, thereby improving the chances of predicting collisions of space objects. The data released from this research is used to design more advanced DNN model capable of predicting other orbital events as well.

1. **INTRODUCTION**

Previous research has chosen that flight data trained with DNN can help control flight vehicles remotely.[1] GN&C is critical for the launch of spacecraft and can be designed according to the mission, while the flight input parameters can help design an algorithm for the optimization of the trajectory.[2][3] Deep Neural Networks have made remarkable progress and are now more compact and applicable even in the on-board domain, but to minimize computational complexity and memory arrangements, Neural Networks have now reached a state where they can be incorporated into the control mechanism of complex systems such as space vehicles. A nonlinear flight model can be used for path optimization as nominal paths which can be robust, [4] and the use of intelligent control architecture to improve space engine performance.[5] Optimizing the use of resources, such as expensive fuel transported, can be achieved by harnessing a dedicated neural network to provide an optimal trajectory map to avoid collisions which can provide invaluable input to the control system. Many international collaborations are going on, to improve space security and ensure long-term space sustainability, including collision avoidance, planetary defense, space debris mitigation, etc.[6] Using the defined data, DNNs are trained to meet the need for trajectory optimization.[7] The collision avoidance of space objects is an overlay on the desired trajectory intended for a mission and, therefore, an integrated neural network obtained through an input based on the control theory, can be used in order to provide the neural network with data from the sensor in real-time and also on the ground on analysis data on the tracking of space objects to allow the DNN to design a prudent trajectory with optimal use of fuel and also prevention of collisions. More than ten years of collaboration between the FGAN and the European Space Operations Center (ESOC) have marked collision avoidance projects.[8] The trained DNNs are used in interplanetary missions to optimize the trajectories of spacecraft to reach the target.[9] Adding the neural network (NN) to the existing trajectory control system is a self-organized function routine. The numerical analysis made it possible to verify the performances given by a specific model.[10] The results of experiments between control data and ML features helped to obtain a clear and comparable performance of the spacecraft.[11]

## 2. MISSION ANALYSIS

*Dikpola-1 Mission* consists of launching five satellites, including three in GEO *(DikpolaSat-1, DikpolaSat-2, DikpolaSat-3)* and the other two in LEO *(DikpolaSat-4 and DikpolaSat-5)*. *Dikpola* is the name given to the satellite mission and it means "The Overseer". Instead of launching five satellites in LEO or in GEO, the mission proposed to launch three satellites in GEO with the same configuration as the other two satellites placed in LEO. This minimizes losses and errors since all the five satellites are made of the same technology and can communicate with each other in a better way than usual. *DikpolaSat-5*'s mission aims to track and de-orbit space debris that will be burnt in Earth orbit. However, in this article we will not provide details of *DikpolaSat-5*, instead, we will focus on *DikpolaSat-4* which mission is to achieve collision avoidance through a trained DNN used for trajectory optimization and control of the trajectory. Each *DikpolaSat* has integrated subsystems including TCS, ACDS, electronic and communication systems, power systems (batteries and solar panels), mobility and deployment mechanism, integrated propulsion systems, and optoelectronic systems.

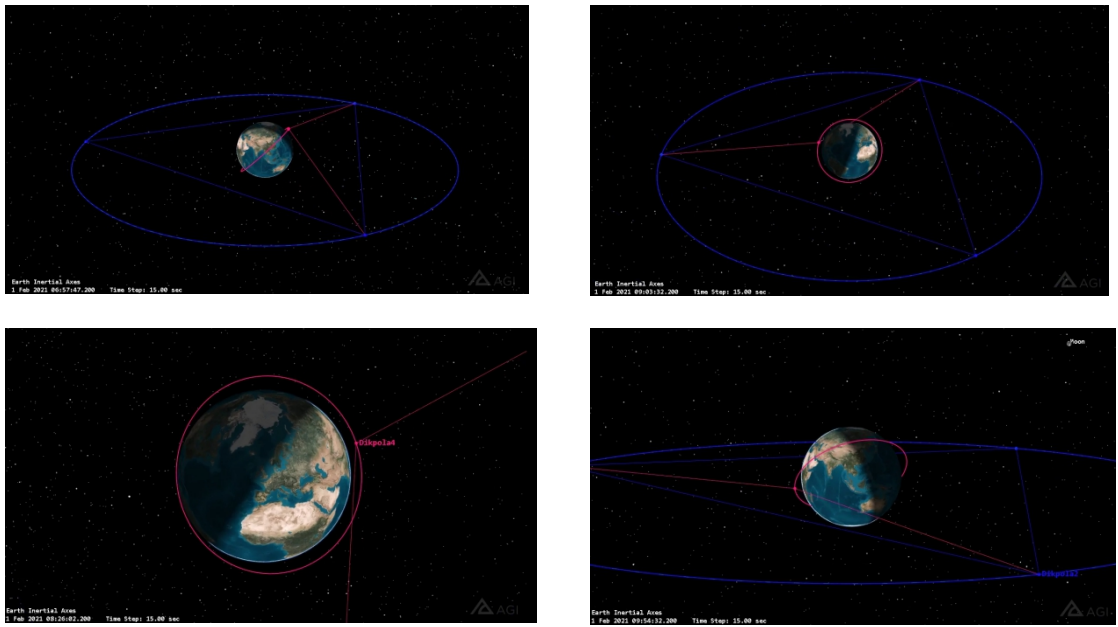

**Figure 1. Digital Mission Analysis of DikpolaSat Mission.** Three satellites in GEO monitoring space objects in LEO, covering an area of 120 degrees each (120 * 3 = 360 degrees). The integrated communication systems available in every satellite, allow them to communicate between and transfer the collected information to *DikpolaSat-4* in LEO. The data transferred to *DikpolaSat-4* in LEO, are analyzed and processed to train a DNN model implemented in the on-board computer of *DikpolaSat-4*, for better detection of spatial objects. The data provided by *DikpolaSat-1,2 and 3*, while monitoring the earth's surface, is processed to create a deep neural network model for collision prediction and avoidance. After detecting the objects, the trained DNN acts on the spacecraft's sensors to activate the path change through the trajectory parameters to avoid collision with space debris, satellites, or others space objects; based on their trajectory and their speed.

### 3. METHODS AND RESULTS

Obstacle avoidance control needs to be redesigned for each new vehicle in flight, as each system has its limits defined very specifically from the maximum thrust that can be provided by the engine used, to the precision with which the attitude system can respond to. Trajectory optimization using a neural network to control the adaptive nature of the nonlinear mechanisms in the controller will make the control system capable of handling multiple nonlinear events and also uncertainties that have not been induced in the control algorithm. By acting on the internal detectors and sensors of the spacecraft, the optimization of resources such as energy, the expensive fuel transported, can be achieved by implementing trained deep neural networks. The DNNs will provide an optimal trajectory control map with invaluable input for collision avoidance.

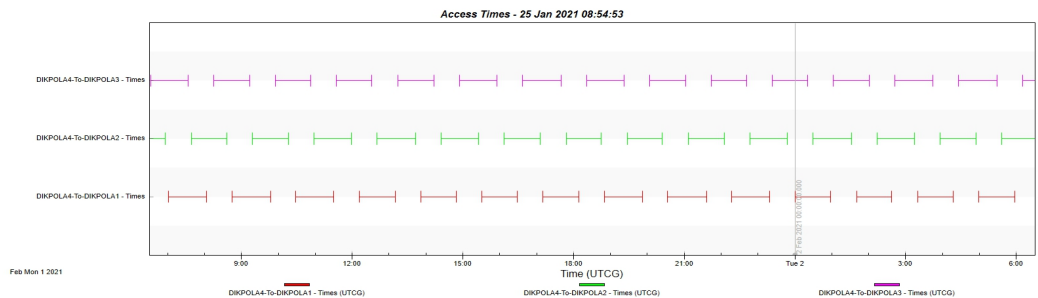

**Figure 2. Regular coverage and data transmission of DikpolaSat Mission.**

The number of variables and constraints in a DNN model also increases with simple shooting implementations. The implementation of DNN with the multifunctional method for undefined symmetric matrices leads to an efficient resolution of the system. In general, all other problems associated with just direct and indirect shooting still apply. The difficulty with shooting methods is the need to define outputs before training the DNNs. DNNs are represented to learn the functional relationship and predict flight states as well as the optimal actions taken. The basic elements used in specifying a model nonetheless, using a homotopy method. From data collection and preparation, selection and training of DNN models to solving the path optimization problem, it has been shown that there is a wide range of possible ways to put the pieces together for the full algorithm. It is necessary to describe the procedure for evaluating DNN models for each method and algorithm. The most successful direct shooting applications have one characteristic in common, namely the ability to describe the problem in terms of a relatively small number of entries.

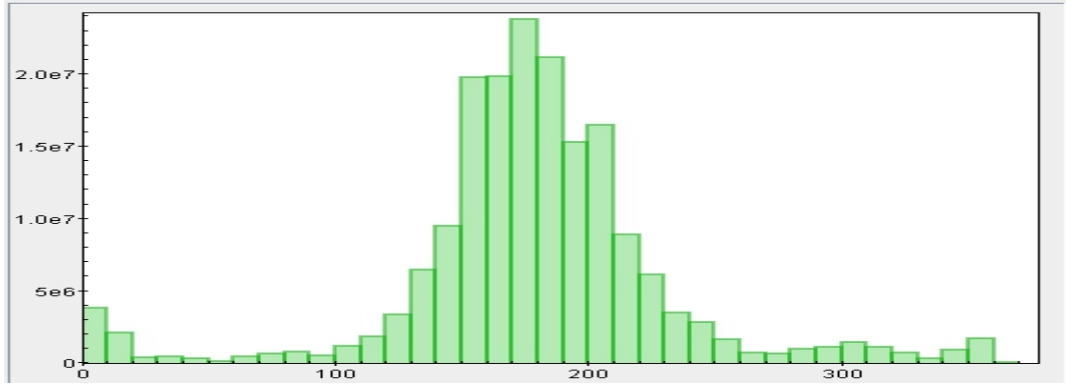

**Figure 3. Constrained Trajectory Optimization of inclination parameters.** A constrained accuracy of [150 ; 210] per unit inclination.

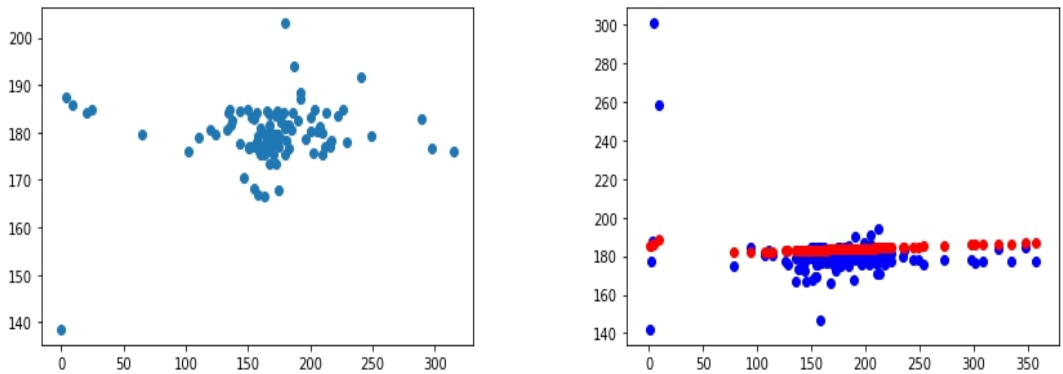

**Figure 4. Optimization of inclination parameters using Deep Neural Networks.**

An offline trained neural network which can provide adaptive tuning parameters for the linear control block and also function as an input for the adaptive nonlinear control system. This is a direct approach similar to the generation. The nonlinear adaptive control system can be integrated inside the neural network itself keeping in mind that neural networks are nonlinear in their true sense, some specific activation functions themselves may vary well to model the non-linearities and the back propagation can manage the error in the dynamics.

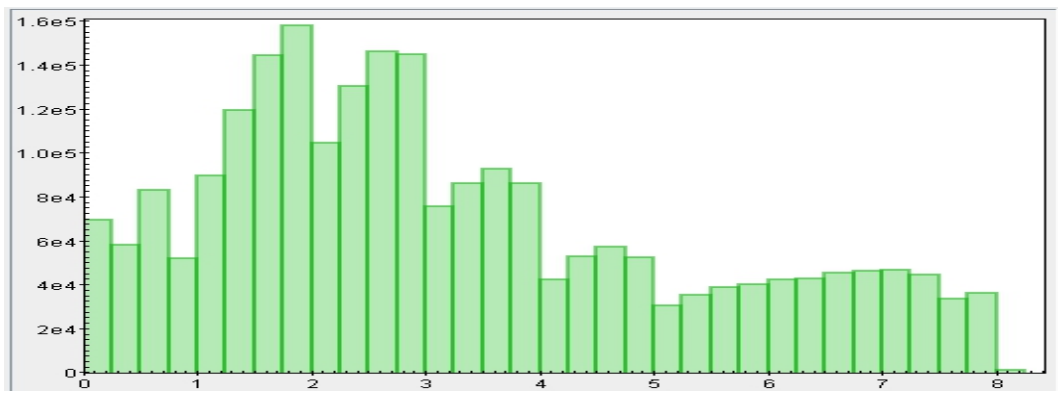

**Figure 5. Constrained Trajectory Optimization of altitude parameters.** A constrained accuracy of [1.25 ; 3] per unit altitude.

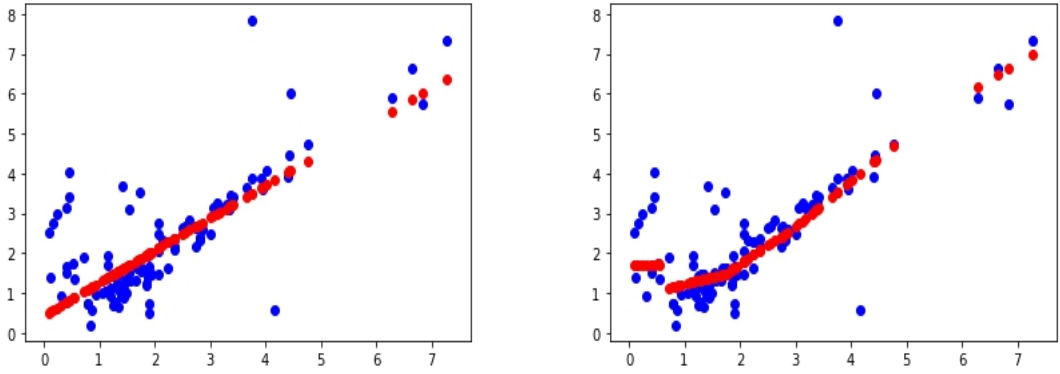

**Figure 6. Optimization of altitude parameters using Deep Neural Networks.**

A whole new variety of neural network can be set up which has 3 different outputs, each neural network having an input channel and added in a deep network that performs the training and gives optimum control adaptability to the control system. Each function generator uses path parameters as inputs and constraints as outputs. We considered three main parameters as inputs: inclination, altitude and velocity. From these analysis, it has been shown that using the DNN algorithms, the optimization of the trajectory parameters can be achieved with high level precision.

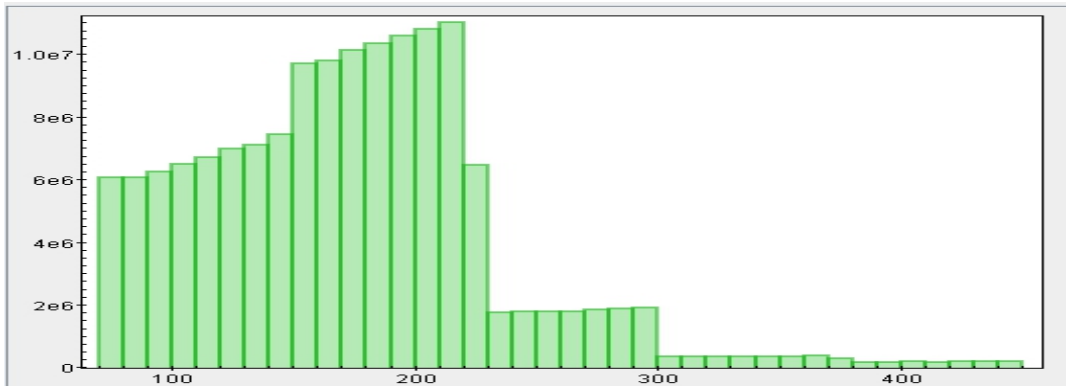

**Figure 7. Constrained Trajectory Optimization of velocity parameters.** A constrained accuracy of [150 ; 220] per unit velocity.

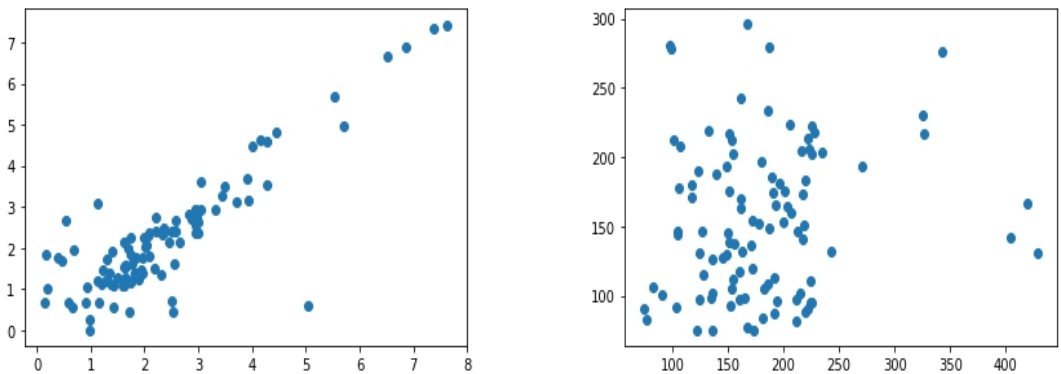

**Figure 8. Optimization of velocity parameters using Deep Neural Networks.**

## 4. CONCLUSION

The limited number of parameters and data can degrade the success of a method. This could be avoided by implementing a DNN algorithm to allow the spacecraft to predict the collisions. The accuracy of the gradient information is also a problem to be solved. DNN implementations use forward difference estimates until they are nearly convergent, then switch to more precise derivatives for convergence. Although finite difference perturbation size selection techniques may appear essential for accurate gradient assessment, a number of efficient methods are available to monitor the interaction between DNN algorithms. The sophisticated algorithms of DNN also make it possible to reduce or even cancel the losses in the gradients. Thus the DNNs improve both the efficiency of the integration and the efficiency of the optimization. The main difficulty with these methods is getting started; that is, find a first estimate of the unspecified conditions at one end that produces a solution reasonably close to the conditions specified at the other end. The reason for this particular difficulty is that extreme solutions are often very sensitive to small changes in unspecified boundary conditions.

**Abbreviations**

DNN - Deep Neural Network

GN&C - Guidance, Navigation & Control

ML - Machine Learning

GEO - Geosynchronous Orbit

LEO - Low Earth Orbit

NLP - Nonlinear Programming

8) Detecting, Tracking And Imaging Space Debris D. Mehrholz Et Al. ESA Bulletin 109 — February 2002

9) Real-Time Guidance For Low-Thrust Transfers Using Deep Neural Networks. Dario Izzo And Ekin Ozturk. 10.2514/1.G005254

10) A New Guidance Algorithm Against High-Speed Maneuvering Target. Tae-Hun Kim. 10.1007/s42405-020-00347-7

11) Deep Reinforcement Learning For Spacecraft Proximity Operations Guidance. Kirk Hovell And Steve Ulrich. 10.2514/1.A34838